\documentclass[runningheads]{llncs}
\usepackage{graphicx}
\usepackage{amsmath}
\usepackage{amssymb}
\usepackage{nccmath}
\usepackage{caption}
\usepackage{multirow}
\usepackage{booktabs}
\usepackage{xcolor}
\begin{document}
\title{{\large{Unsupervised Deformable Registration for Multi-Modal Images via Disentangled Representations}}}   

%
\titlerunning{To be published in the proceedings of IPMI 2019}
\author{Chen Qin\inst{1,2}\thanks{This work was conducted when Chen Qin was an intern AI scientist at Siemens Healthineers.} \and
Bibo Shi\inst{2} \and
Rui Liao\inst{2} \and Tommaso Mansi\inst{2} \and Daniel Rueckert\inst{1} \and Ali Kamen\inst{2}}
%
\authorrunning{To be published in the proceedings of IPMI 2019}
\institute{Department of Computing, Imperial College London, London, UK \\
\email{c.qin15@imperial.ac.uk} \and
Digital Services, Digital Technology \& Innovation, Siemens Healthineers, Princeton, New Jersey, USA\\ 
\email{bibo.shi@siemens-healthineers.com} }

\maketitle              
\setcounter{footnote}{0}

\begin{abstract}
We propose a fully unsupervised multi-modal deformable image registration method (UMDIR), which does not require any ground truth deformation fields or any aligned multi-modal image pairs during training. Multi-modal registration is a key problem in many medical image analysis applications. It is very challenging due to complicated and unknown relationships between different modalities. In this paper, we propose an unsupervised learning approach to reduce the multi-modal registration problem to a mono-modal one through image disentangling. In particular, we decompose images of both modalities into a common latent shape space and separate latent appearance spaces via an unsupervised multi-modal image-to-image translation approach. The proposed registration approach is then built on the factorized latent shape code, with the assumption that the intrinsic shape deformation existing in original image domain is preserved in this latent space. Specifically, two metrics have been proposed for training the proposed network: a latent similarity metric defined in the common shape space and a learning-based image similarity metric based on an adversarial loss. We examined different variations of our proposed approach and compared them with conventional state-of-the-art multi-modal registration methods. Results show that our proposed methods {{achieve competitive performance against other methods at substantially reduced computation time}}.
\end{abstract}
\section{Introduction}
Different medical image modalities, such as Magnetic Resonance Imaging (MRI), Computed Tomography (CT), and Positron Emission Tomography (PET), show unique tissue features at different spatial resolutions. In clinical practice, multiple image modalities must often be fused for diagnostic or interventional purpose, providing the combination of complementary information. However, images from different modalities are often acquired with different scanners and at different time points with some intra-patient anatomical changes. It is of great importance to register multi-modal images for an accurate analysis and interpretation. 

Multi-modal image registration is a challenging problem, due to the unknown and complex relationship between the intensity distributions of the images to be aligned. Also, there could be presence of features in one modality but missing in another. Previous multi-modal image approaches either rely on information theoretic measures such as mutual information or on landmarks being identified in both images. However, information theoretic measures often ignores spatial information, and anatomical landmarks cannot always be localized in both images. Furthermore, landmark detection  can be time-consuming or impossible in image-guided intervention.

In this paper, we propose a novel unsupervised registration method for aligning {{intra-subject}} multi-modal images, without the need of ground truth deformation fields, aligned multi-modal image pairs or any anatomical landmarks during training. To address this, our main idea is to learn a parameterized registration function via reducing the multi-modal registration problem to a mono-modal one in latent embedding space. In particular, our method decomposes images into a domain-invariant latent shape representation and a domain-specific appearance code based on the multi-modal unsupervised image-to-image translation framework (MUNIT) \cite{huang2018multimodal}. With the assumption that the intrinsic shape deformation between multi-modal image pairs is preserved in the domain-invariant shape space, we propose to learn an unsupervised diffeomorphic registration network directly based on the disentangled shape representations. A similarity criterion thus can be defined in the latent space, minimizing the latent shape distance between warped moving image and target one. Additionally, a complimentary learning-based similarity metric is also proposed, which is defined via an adversarial loss to distinguish whether a pair of images are sufficiently aligned or not in the image domain. Since transformation is learned from a domain-invariant space, the method is directly applicable to bi-directional multi-modal registration without extra efforts.

Our main contributions can be summarized as follows: First, we present a learning-based unsupervised multi-modal deformable image registration method that does not require any aligned image pairs or anatomical landmarks. Second, we propose to learn a {\it{\textbf{bi-directional}}} registration function based on disentangled shape representation by optimizing the proposed similarity criterion defined on both latent and image space. Third, we demonstrate that our proposed methods are competitive to state-of-the-art multi-modal image registration solutions in terms of accuracy, and have a much faster speed. To the best of our knowledge, this is the first work investigating a {\it{\textbf{fully unsupervised}}} deep learning based method for {\it{\textbf{multi-modal}}} deformable image registration. Though our work is currently demonstrated on 2D images, it can be readily extended for 3D volumes.

{\textbf{Related Work:}} One category of the classical and standard methods for multi-modal registration are information theory based approaches, which utilize mutual information (MI) as a similarity measure to align multi-modal images. It showed a great success in rigid registration of multi-modal medical images, and later its variations such as normalized MI and local MI etc. \cite{pluim2003mutual} have also been proposed to tackle deformable registration. However, such methods are often based on intensity probability distribution, and thus ignore spatial information of anatomical structures. 
An alternative way to address multi-modal image registration problem is to reduce the problem to a mono-modal one. They either synthesize one modality from another or map both modalities to a common domain. In order to reduce the appearance gap between different modalities, image synthesis can be achieved by taking advantage of prior knowledge on physical properties of imaging devices \cite{roche2001rigid} or capturing intensity relationships using learning-based methods \cite{cao2017dual}. As to mapping both modalities to a common space, the assumption is that both modalities share the same anatomical structure or feature, and thus can be used to establish meaningful correspondences. 

In recent years, many deep learning approaches have also been proposed in image registration domain. In supervised setting, these methods require ground truth deformation fields during the training process. However, as ground truth deformations are rarely available, they commonly synthetically generate geometric deformations as ground truth and then transform one of the image pairs \cite{ilg2017flownet,uzunova2017training}. Some other methods employ kind of weakly-supervised way for image registration, where they rely on the alignment of multiple labelled corresponding anatomical structures for individual image pairs during the training \cite{hu2018weakly}. On the other hand, for unsupervised deformable image registration, most current approaches proposed to use convolutional neural networks \cite{balakrishnan2018unsupervised,qin2018joint} or probabilistic framework \cite{dalca2018unsupervised,krebs2018unsupervised} with a spatial transformation function \cite{jaderberg2015spatial}, which were trained by minimizing conventional image similarity metrics.
Instead of using specific similarity metrics, Fan et al. \cite{fan2018adversarial} proposed a similar registration network which was trained along with learning a similarity measure by using a discriminator network to judge whether a pair of images are sufficiently aligned. However, these unsupervised methods mainly focus on mono-modal image registration. 

\section{Proposed Method}
\label{Method}
Our goal is to learn a multi-modal deformable registration network in a fully unsupervised manner: without ground truth deformation fields, anatomical landmarks, or aligned multi-modal image pairs for training. We achieve this by embedding images of different modalities into a domain-invariant space via image disentangling, where any meaningful geometrical deformation can be directly derived in the latent space. 
Our method mainly consists of three parts: image disentangling network via unpaired image-to-image translation, a deformable registration network in the disentangled latent space and an adversarial network that implicitly learns a similarity metric in image space. A schematic illustration of our method is shown in Fig. \ref{fig:munit} and Fig. \ref{fig:UMDIR}.

\subsection{Image disentangling via unpaired image-to-image translation}
\label{image_translation}
\begin{figure*}[!t]
\centering
\includegraphics[width=0.95\linewidth]{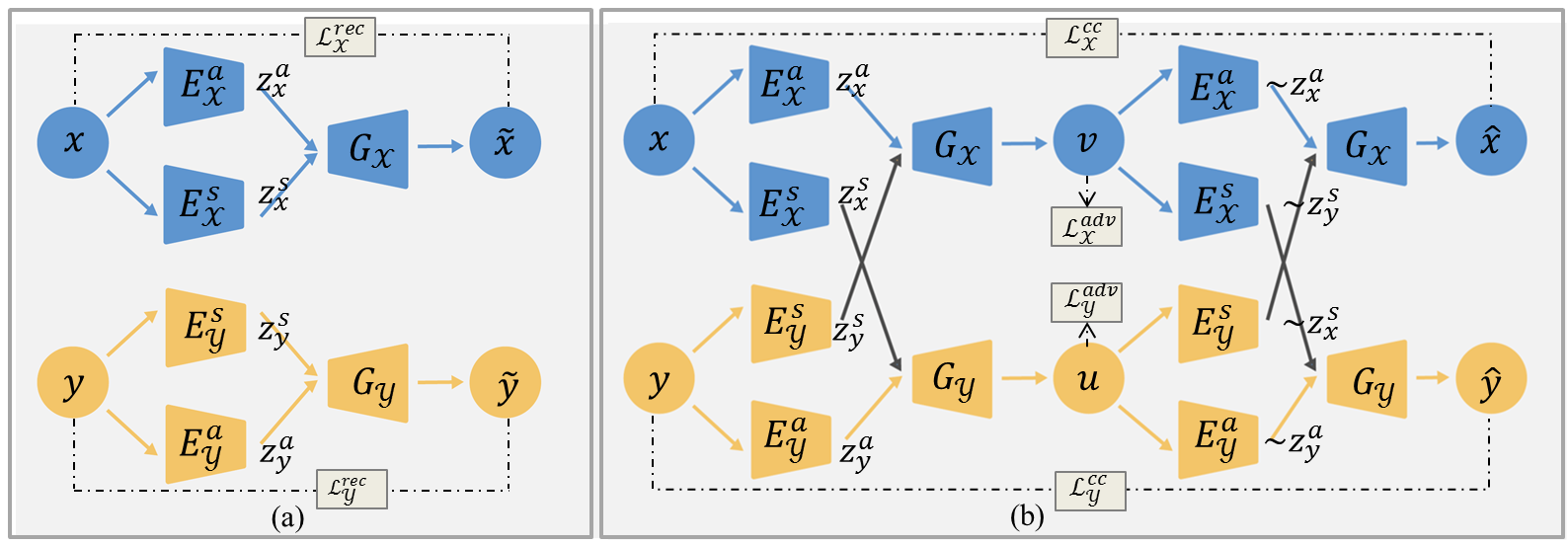}
\caption{Image-to-image translation framework (Section \ref{image_translation}). $x$ and $y$ are sample images in $\mathcal{X}$ and $\mathcal{Y}$ domain respectively. $\left\{E_\mathcal{X}^s, E_\mathcal{Y}^s\right\}$ and $\left\{E_\mathcal{X}^a, E_\mathcal{Y}^a\right\}$ are corresponding shape encoders and appearance encoders. $\{G_{\mathcal{X}}, G_{\mathcal{Y}}\}$ are image generators. (a) Image self-reconstruction with $\{\tilde{x}, \tilde{y}\}$ reconstructed from $\{x, y\}$. (b) Image{{-to-image}} translation and cross-cycle reconstruction. $\{u, v\}$ are translated images from $\{x, y\}$ to domain $\{\mathcal{Y}, \mathcal{X}\}$ respectively. $\{\hat{x}, \hat{y}\}$ are cross-cycle reconstructed images. Black dotted lines {{and tilde notation }}indicate the consistency between variables.}
\label{fig:munit}
\end{figure*}
Huang et al. \cite{huang2018multimodal} and Lee et al. \cite{lee2018diverse} have proposed to solve unpaired image-to-image translation problem through disentangled image representations, 
{{where images are embedded into a domain-invariant attribute space and a domain-specific attribute space, as shown in Fig.\ref{fig:munit}. As described and shown in \cite{huang2018multimodal}, domain-invariant attribute mainly captures the underlying spatial structure, and domain-specific attribute corresponds to the rendering of structure that is determined by imaging physics in our application.}}
This approach formed the basis of our work. We briefly describe its main concept below that is related to our following registration work. 

Let $x \in \mathcal{X}$ and $y \in \mathcal{Y}$ denote unpaired images from two different {{domains, or in our application, two different imaging modalities}}. As illustrated in Fig.\ref{fig:munit}, image $x$ is disentangled into a shape (content) code $z_x^s$ in a domain-invariant space $\mathcal{S}$ and an appearance code $z_x^a$ in a domain specific space $\mathcal{A}_{\mathcal{X}}$, where $E_\mathcal{X}^s$ and $E_\mathcal{X}^a$ encode $x$ to $z_x^s$ and $z_x^a$ respectively. The generator $G_{\mathcal{X}}$ generates images conditioned on both shape and appearance vectors. Image-to-image translation is performed by swapping the latent codes in two domains, such as $v=G_{\mathcal{X}}(z_x^a, z_y^s)$ so that image $y$ is translated to target domain $\mathcal{X}$. To train the framework for image{{-to-image}} translation and achieve representation disentanglement, a bidirectional reconstruction loss is used which comprises image self-reconstruction loss and latent reconstruction loss, i.e.,
\begin{subequations}
\small
\begin{align}
 \mathcal{L}_{\mathcal{X}}^{rec} &= \mathbb{E}_x\big[||G_{\mathcal{X}}(E_\mathcal{X}^s(x), E_\mathcal{X}^a(x))-x||_1\big],\\
    \mathcal{L}_{\mathcal{X}^s}^{lat} &= \mathbb{E}_{x,y}\big[||E_{\mathcal{Y}}^s(G_{\mathcal{Y}}(z^s_x, z^a_y))-z^s_x||_1\big],
\\
    \mathcal{L}_{\mathcal{Y}^a}^{lat} &= \mathbb{E}_{x,y}\big[||E_{\mathcal{Y}}^a(G_{\mathcal{Y}}(z^s_x, z^a_y))-z^a_y||_1\big].
    \end{align}
\end{subequations}
In order to better preserve the shape information, we also propose to incorporate an extra loss term to ensure cross-cycle consistency \cite{lee2018diverse} :
\begin{equation}
\resizebox{.92\hsize}{!}{$
\begin{aligned}
    \mathcal{L}^{cc} &= \mathcal{L}_{\mathcal{X}}^{cc} + \mathcal{L}_{\mathcal{Y}}^{cc}= \mathbb{E}_{x,y}\big[||G_{\mathcal{X}}(E_\mathcal{Y}^s(u), E_\mathcal{X}^a(v))-x||_1 +||G_{\mathcal{Y}}(E_\mathcal{X}^s(v), E_\mathcal{Y}^a(u))-y||_1\big].
\end{aligned}$}%
\end{equation}
Besides, adversarial losses $L_{\mathcal{X}}^{adv}$ and $L_{\mathcal{Y}}^{adv}$ are employed to match the distribution of translated images to the image distribution in the target domain. Overall, our image{{-to-image}} translation network is trained by a weighted sum of image self-reconstruction loss, latent representation reconstruction loss, adversarial loss and the cross-cycle consistency loss. For more details, please refer to \cite{huang2018multimodal,lee2018diverse}.

\begin{figure*}[!t]
\centering
\includegraphics[width=0.95\linewidth]{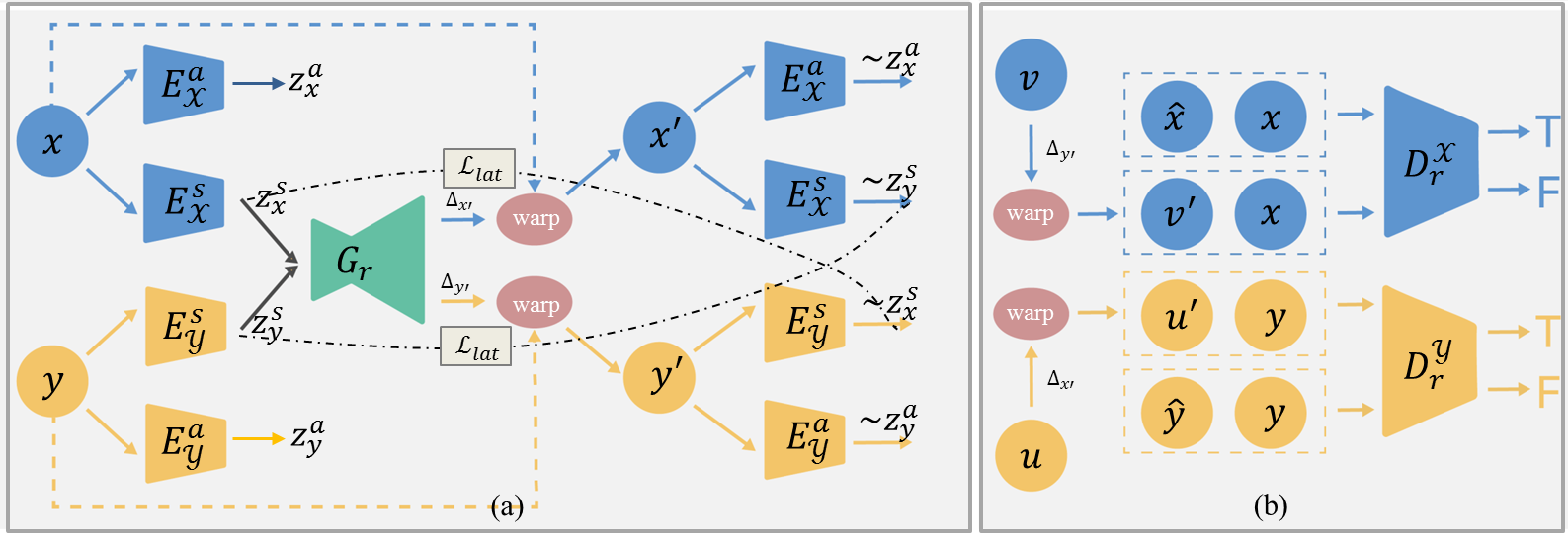}
 \caption{Overview architecture of the proposed models. (a) Multi-modal image registration via disentangled representations (Section \ref{latent_space}). {{$x'$ and $y'$ are warped images from $x$ and $y$.}} (b) Learning-based similarity metric in image space. $u$, $v$ {{and $\hat{x}$, $\hat{y}$}} are translated {{and reconstructed images respectively}} adopted from Fig.\ref{fig:munit}. (a)+(b) Multi-modal image registration via combined metrics (Section \ref{combined_space}).  $G_{r}$ is the registration network in latent space. $D_{r}^{\mathcal{X}}, D_{r}^{\mathcal{Y}}$ are discriminators in image space. }
\vspace{-0.1in}
\label{fig:UMDIR}
\end{figure*}

\subsection{Multi-modal image registration via disentangled representations}
\label{latent_space}
{{With image{{-to-image}} translation and disentangled attributes being learned, we are able to reduce multi-modal registration problem to a mono-modal one by embedding images onto the domain-invariant latent space and learn the deformation there.}} A explanatory figure of our proposed network is shown in Fig.\ref{fig:UMDIR}(a). 

Specifically, images from different modalities are disentangled into a shared shape space $\mathcal{S}$ and different appearance spaces $\mathcal{A}_{\mathcal{X}}$ and $\mathcal{A}_{\mathcal{Y}}$ respectively. The latent shape representations $z_x^s$ and $z_y^s$ contain high-level structure information of images which is capable of restoring the original image by combining with the appearance code. Relying on this, we propose to learn a deformable registration network by aligning images via these disentangled shape representations. When registering a moving image $y \in \mathcal{Y}$ to a fixed image $x \in \mathcal{X}$, the structure of the warped moving image $y' \in \mathcal{Y}$ should be close to that of the fixed one while keeping the appearance unchanged. Therefore, a similarity criterion for training the registration network can be defined in the disentangled latent shape space.
Specifically, we propose to learn a diffeomorphic registration network that receives latent shape representations as inputs and predicts a velocity field $w$. Deformation  $\Delta$ between moving and fixed images is defined as an exponential map with respect to the velocities: $\Delta = \text{exp}(w)$, which is implemented by an exponentiation layer as proposed in \cite{krebs2018unsupervised}. The detailed architecture of the registration network follows an idea originally proposed in \cite{qin2018joint}. 
To train the network, the warped image $y'$ is then encoded back to the latent shape space, and thus similarity between shape representations $E_\mathcal{Y}^s\left(y'\right)$ and $z_x^s$ can be enforced. In addition, since both images are mapped to a common feature space (modality-independent space), the registration network learned in this space is directly applicable to be {\it{\textbf{bi-directional}}}, {{i.e., for both $y \rightarrow x$ and $x \rightarrow y$ registration}}. This is superior to learning a registration network in image space, which normally requires separate training for each direction. 
Therefore, by incorporating the bi-directional registration, the network can be trained by minimizing the following similarity metric that is defined on latent space:
\begin{equation}
\resizebox{.9\hsize}{!}{$
\begin{aligned}
\mathcal{L}_{lat} = \mathbb{E}_{x,y}\big[||E_\mathcal{Y}^s\left(y'\right)-z_x^s||_1+||E_\mathcal{X}^s\left(x'\right)-z_y^s||_1\big] +\lambda_{\Delta} \big[\mathcal{H}(\triangledown_{i,j}\Delta_{y'})+\mathcal{H}(\triangledown_{i,j}\Delta_{x'})\big],
\end{aligned}$}%
\end{equation}
where we penalize the gradients of the deformation fields $\Delta_{y'}$ and $\Delta_{x'}$ using an approximation of Huber loss \cite{qin2018joint} {\small$\mathcal{H}(\triangledown_{i,j}\Delta)= \sqrt{\epsilon+\sum_{m=i,j}(\triangledown_{i}\Delta m^2+\triangledown_{j}\Delta m^2)}$} along both $i$ and $j$ directions to ensure the smoothness. $\lambda_{\Delta}$ is a regularization parameter for a balance (trade-off) between different terms, and $\epsilon=0.01$.

\subsection{Multi-modal image registration via combined metrics}
\label{combined_space}
While disentangled latent shape representations can effectively capture high-level structural information, training with a latent similarity criterion only could possibly ignore some detailed structure deformations. To compensate this, we propose to combine the latent similarity criterion with an additional learning-based similarity metric in image space, as shown in Fig.\ref{fig:UMDIR}(b). 

Similarly, here we define the learning-based similarity metric in image space also via image{{-to-image}} translation. However, during image-to-image translation, there could inevitably exist some mismatch of distributions between synthesized images and target images, especially when appearance distributions of real images are complex. Thus, mono-modal registration methods based on intensity similarities may not be sufficient. Therefore, instead of using a specific intensity-based similarity measure, similar to \cite{fan2018adversarial}, we propose to learn a similarity metric function formulated by a patch GAN discriminator, which is trained to distinguish if a pair of image patches is well-aligned or not. Different from \cite{fan2018adversarial}, to mitigate influence of distribution mismatch, we utilize the cross-cycle consistency of the translation network when designing the real pairs (well-aligned) and fake pairs (registered by network), i.e., {\small  $\left\{G_\mathcal{X}\left(E_\mathcal{Y}^s(u), E_\mathcal{X}^a(v)\right), x \right\}$} and $\left\{v', x \right\}$, where $v'$ indicates the corresponding warped images of $v$. This is to enforce the discriminator to learn structure alignment instead of distribution differences. Architecture of discriminators follows the design of the feature encoder in registration network. Overall, we formulate the combined problem using the improved Wasserstein GAN (WGAN-GP)\cite{gulrajani2017improved}: the image registration network $G_{r}$ (generator) and two discriminators $D_{r}^{\mathcal{X}}$ and $D_{r}^{\mathcal{Y}}$ can be trained via alternatively optimizing the respective composite loss functions:
\begin{subequations}
\small
\begin{align}
\mathcal{L}_{D_{r}^{\mathcal{X}}} &= {\mathbb{E}}_{\tilde{q} \sim \mathbb{P}_{f}}\big[D_{r}^{\mathcal{X}}(\tilde{q})\big]-{\mathbb{E}}_{{q} \sim \mathbb{P}_{r}}\big[D_{r}^{\mathcal{X}}({q})\big]+\lambda_{grad}\cdot\mathcal{L}_{grad}^{\mathcal{X}}
\\ 
\mathcal{L}_{D_{r}^{\mathcal{Y}}} &= {\mathbb{E}}_{\tilde{p} \sim \mathbb{P}_{f}}\big[D_{r}^{\mathcal{Y}}(\tilde{p})\big]-{\mathbb{E}}_{{p} \sim \mathbb{P}_{r}}\big[D_{r}^{\mathcal{Y}}({p})\big]+\lambda_{grad}\cdot\mathcal{L}_{grad}^{\mathcal{Y}}
\\ 
\mathcal{L}_{G_{r}} &= -{\mathbb{E}}_{\tilde{q} \sim \mathbb{P}_{f}}\big[D_{r}^{\mathcal{X}}(\tilde{q})\big]-{\mathbb{E}}_{\tilde{p} \sim \mathbb{P}_{f}}\big[D_{r}^{\mathcal{Y}}(\tilde{p})\big] + \alpha \mathcal{L}_{lat},
\label{eq:data_fidelity}
\end{align}
\end{subequations}
where $D_{r}^{\mathcal{X}}$ and $D_{r}^{\mathcal{Y}}$ are two discriminators for the bi-directional registration to distinguish real pairs and fake pairs in $\mathcal{X}$ and $\mathcal{Y}$ domain. \{$q, \tilde{q}$\} and \{$p, \tilde{p}$\} are \{real, fake\} pairs sampled from $\mathcal{X}$ and $\mathcal{Y}$ respectively. {\small$\mathcal{L}_{grad}^{\mathcal{X}}$} is the gradient penalty for the discriminator $D_{r}^{\mathcal{X}}$ which can be expressed as the form of {\small$\mathcal{L}_{grad}^{\mathcal{X}}={\mathbb{E}}_{\hat{q} \sim \mathbb{P}_{\hat{q}}}\big[\left(||\triangledown_{\hat{q}}D_{r}^{\mathcal{X}}(\hat{q})||_2-1\right)^2\big]$} with $\hat{q}$ sampled uniformly between $q$ and $\tilde{q}$, and the same with {\small$\mathcal{L}_{grad}^{\mathcal{Y}}$}. $\alpha$ is a parameter to balance between the learning-based image space similarity metric and the latent space similarity measure.

\section{Experiments}

\subsection{Datasets}
We used two datasets for evaluation: one with clinical meaningful deformations in single modality (COPDGene) and one with real well-aligned multi-modality images (BraTS). In this case, we can have control over image-to-image translation quality and predicted deformation respectively.  

{\bf{COPDGene}}\footnote{The COPDGene study (NCT00608764) was funded by NHLBI U01 HL089897 and U01 HL089856 and also supported by the COPD Foundation through contributions made to an Industry Advisory Committee comprised of AstraZeneca, Boehringer-Ingelheim, GlaxoSmithKline, Novartis, and Sunovion.} 
The COPDGene study is a multicenter observational study to analyze genetic susceptibility for the development of chronic obstructive pulmonary disease (COPD)\cite{regan2011COPD}. High-quality, volumetric lung CT scans were acquired to capture full inspiration cycle of each subject using a standardized imaging protocol. {CT scans were reconstructed with slice thicknesses of 0.625, 0.75, or 0.9 mm depending on the CT scanner manufacturer, with corresponding slice intervals of 0.625, 0.5, and 0.45 mm, respectively.} 
In our experiment,  1000 subjects are randomly retrieved for evaluation, with end inspiration and expiration volumes being used to derive the underlying breathing motion. Each pair of volumes was rigidly pre-aligned, cropped, and down-sampled into a 3D volume with size of $128\times128\times128$ and resolution of 2.5mm. We randomly split the 1000 subjects into 800/100/100 for train/test/validation, and on each subject we extract middle 10 slices. 
To simulate a multi-modality image registration problem, we synthesized a new modality using an intensity transformation $cos(I \cdot \pi)$ as proposed in \cite{yaman2015iterative} followed by Gaussian blurring and intensity normalization. Deformations were estimated between end inspiration and expiration frames of real CT and synthesized images. 

{\bf{BraTS'17}} 
The Brain Tumour Segmentation (BraTS) 2017 dataset is obtained from the MICCAI BraTS 2017 challenge \cite{menze2015multimodal,bakas2017advancing}. Specifically, it provides a large dataset of multi-modal MRI scans (native T1, T2, T2-FLAIR, and T1Gd) for patients with glioblastomas. Overall, the available training set consists of 285 cases, and for each case four image modalities were standardized into a 3D volume in size of $240 \times 240 \times 155$ with 1 mm isotropic resolution. 
In our experiments, we utilize the T1 and T2-weighted images to define a multi-modality dataset to demonstrate the effect of our proposed approach. The set is randomly split into 225/30/30 for train/validation/test, and central 20 slices of each subject were extracted. 
As provided T1 and T2 images are already aligned, we generated synthetic deformation fields by spatially transforming one of the modality (T1) using elastic transformations on control points followed by Gaussian smoothing. The synthetic deformation is only used as ground truth (GT) for evaluations.

\vspace{-1.5mm}
\subsection{Experimental Settings}
\textbf{Implementation Details:} For image-to-image translation, we built our network based on MUNIT implementation with changes as discussed in Section \ref{image_translation}. The network is trained using the default settings as in \cite{huang2018multimodal}. {{Our registration network adopts the same architecture as in \cite{qin2018joint} with an additional exponentiation layer \cite{krebs2018unsupervised} as the last layer}}. In our implementation, we pre-train the image-to-image translation network using unpaired images, and then multi-modal registration and discriminator networks are trained. Our networks are implemented on PyTorch, using Adam optimizer 
for training with a learning rate of 0.0001. Hyper-parameters were chosen based on the performance on validation set, with $\lambda_{grad}=10$ and $\lambda_\Delta = 1$. Run time reported for each method was tested on the same PC with 32G RAM, 3.6GHz CPU, and Quadro P4000 GPU.

{\textbf{Evaluation Measures:}} For COPDGene dataset, we evaluate the registration accuracy indirectly via provided lung segmentation masks, as no ground truth deformation fields are available. Dice score, mean contour distance (MCD) and Hausdoff distance (HD) are computed between lung masks of fixed and warped moving images. For BraTS dataset, synthetic deformation fields are used as GT, thus pixel-wise root mean square error (RMSE($\Delta$)) is calculated for the evaluation. Also, as pairs of aligned images are available, pixel-wise intensity error (RMSE(I)) can be calculated when transforming back the deformed image. Additionally, for both datasets, analysis of Jacobian matrix $J_{\Delta}(m)=\triangledown \left(\Delta(m)\right)$ were conducted on the dense deformation fields $\Delta$ of each pixel $m$. Gradients of $J_{\Delta}(m)$ (Grad Det-Jac) are calculated as a metric to show the smoothness of Jacobian. Besides, average run time of each method is reported. 

\textbf{Competing Methods:} We compare with two well-established  multi-modal image registration methods: A mutual information based approach using the \textbf{Elastix} toolbox   \cite{marstal2016simpleelastix} and the \textbf{MIND} approach \cite{heinrich2012mind}. 
Also, we compare with the diffeomorphic Demons method \cite{vercauteren2007non} which can deal with multi-modal images using the proposed learned image-to-image translation network. Specifically, we first translate the appearance of the moving image to that of the fixed image and then run the diffeomorphic Demons algorithm on the image pair.  We term this method as \textbf{I2I+DiffDem}. 
A hierarchical multiresolution optimization scheme was used for all. Parameters were determined via searching on the validation set of each dataset separately while considering both the registration accuracy and speed. Results reported are best performance we can achieve. 
In addition, as no other deep learning based unsupervised multi-modal registration has been proposed yet, here we employ a diffeomorphic extension of existing mono-modal registration method \cite{fan2018adversarial} that is enabled for multi-modal images similarly as I2I+DiffDem, termed as \textbf{I2I+GAN}. Its registration network architecture follows Section \ref{latent_space} where instead of estimating deformation $\Delta$, velocity field is estimated to ensure the diffeomorphism. Its discriminator has also been adapted for multi-modality as discussed in Section \ref{combined_space}. 
Finally, we term variants of our proposed method as \textbf{UMDIR-Lat}, \textbf{UMDIR-GAN} and \textbf{UMDIR-LaGAN}, corresponding to models shown in Fig.\ref{fig:UMDIR} (a), (b) and (a)+(b). 

\subsection{Results}
\begin{table*}[!t]
  \centering
  \caption{Evaluation of multi-modal registration on COPDGene dataset in terms of average Dice score, MCD, HD (unit: pixel), run time (GPU/CPU) and Grad Det-Jac ($\times 10^{-2}$) of each deformation field. }
  \label{COPD_table}
\scalebox{0.92}
 { \begin{tabular}{ccccccc}
  \toprule
Method & Avg. Dice & MCD & HD & Grad Det-Jac& Time(s) \\
  \midrule
    {MIND \cite{heinrich2012mind}} & 0.9365 (0.0822) & 1.168 (0.798) & 11.129 (5.736) & 4.52 (0.40) & -/12.09\\
  {Elastix \cite{marstal2016simpleelastix}} & 0.9497 (0.0822) & 0.934 (0.760) & 9.970 (5.772) & 2.29 (0.61) &  -/105.66\\
  \midrule
   {I2I+DiffDem} & 0.9347 (0.0676) & 1.274 (0.923) & 11.477 (5.434) & 6.14 (0.39) &  2.45/3.96\\
 {I2I+GAN} & 0.9553 (0.0444) & 0.912 (0.628) & 9.383 (4.977) & 2.28 (0.45) & 0.12/2.81\\
 \midrule
UMDIR-GAN & 0.9613 (0.0357) & 0.819 (0.546) & 9.188 (4.981)   &{\bf{2.19}} (0.48) &  \bf{0.07}/1.62 \\
UMDIR-Lat & 0.9603 (0.0349) & 0.823 (0.462) & {{8.469}} (4.491) & 2.73 (0.80)&  \bf{0.07}/1.62\\
 UMDIR-LaGAN & {\bf{0.9672}} (0.0280) & {\bf{0.710}} (0.436)  & {\bf{8.257}} (4.432) & 2.79 (0.60) & \bf{0.07}/1.62\\
    \bottomrule
  \end{tabular}}
\end{table*}

\begin{figure*}[!t]
\centering
\includegraphics[width=0.94\linewidth]{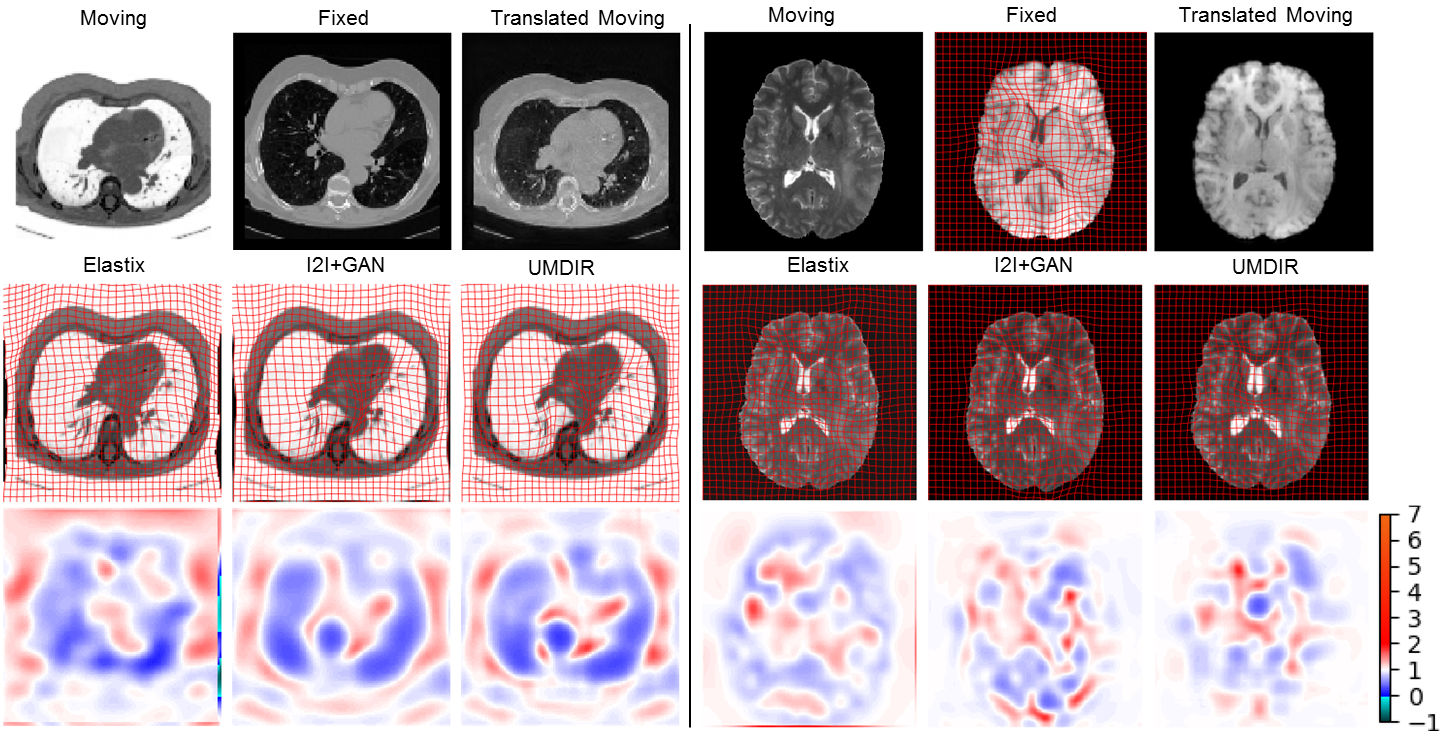}
\caption{Visualization results of our model against baseline methods, where warped moving images, corresponding estimated deformation fields  and Jacobian determinant are shown. Left: COPDGene data; Right: BraTS data with GT overlaid on Fixed image.}
\label{fig:COPD}
\vspace{-3mm}
\end{figure*}

First, we evaluated the performance of our methods on COPDGene data. The quantitative results for this are shown in Table \ref{COPD_table}. The registration performance is evaluated via measuring the Dice score, MCD and HD between warped lung segmentation at expiration and the GT lung segmentation at inspiration, as well as the gradient of Jacobian determinant. It can be seen that compared to the baseline methods, both traditional multi-modal registration methods and image-to-image translation plus mono-modal registration methods, our proposed UMDIR methods outperforms them in terms of Dice, MCD and HD with Grad Det-Jac smaller than or in the same range with other competing methods. {{In particular, success of training with latent similarity criterion implies that the learned domain-invariant attribute is capable of extracting and preserving intrinsic shape feature that is informative enough to guide the geometrical transformation for registration.}} In addition, Fig. \ref{fig:COPD} displays the warped moving images along with their corresponding deformation fields and Jacobian determinant, where it can be observed that our proposed model is able to achieve high accuracy with smooth and regular deformation fields.
Furthermore, by combining both latent similarity and cross-cycle adversarial similarity metrics, we see a further improvement of performance on the COPDGene dataset in terms of accuracy, which indicates that these two metrics could be complementary. In terms of registration speed, our proposed methods are significantly faster than traditional baseline methods. In particular, the UMDIR methods are faster than I2I+GAN, as they learn the registration network directly from latent representations, bypassing the image-to-image translation stage.
Note that the translated moving images in Fig. \ref{fig:COPD} are only used in the I2I+DiffDem and I2I+GAN.

\begin{table*}[!t]
  \centering
  \caption{Evaluation of bi-directional multi-modal registration on BraTS dataset in terms of RMSE($\Delta$) (unit: pixel) and Grad Det-Jac ($\times 10^{-2}$) for T2 $\rightarrow$ T1 registration, and RMSE(I)) for the inverse direction. Average run time (GPU/CPU) is also provided. }
  \label{BraTS_table}
 \scalebox{0.92}
 { \begin{tabular}{cccccc}
  \toprule
\multirow{2}*{Method} & \multicolumn{2}{c}{T2 $\rightarrow$ T1} & \multicolumn{2}{c}{T1 $\rightarrow$ T2}& \multirow{2}*{Time(s)}  \\
\cline{2-5}
 & {RMSE($\Delta$)} & Grad Det-Jac & RMSE(I) & Grad Det-Jac   \\
  \midrule
{MIND \cite{heinrich2012mind}} & 1.266 (0.253) & 3.58 (0.25) & {\bf{0.045}} (0.011) &3.60 (0.24) & -/18.98 \\
  {Elastix \cite{marstal2016simpleelastix}} & 1.260 (0.225) & 1.23 (0.14) & 0.089 (0.013) &  1.22 (0.17) & -/66.97 \\
  \midrule
  {I2I+DiffDem} & 1.391 (0.183) & {\bf{0.83}} (0.13) & 0.057 (0.015) & {\bf{0.87}} (0.14) & 0.81/4.44\\
 {I2I+GAN} & 1.250 (0.218) & 1.30 (0.11) & 0.074 (0.014) & 2.10 (0.43) & 0.20/6.44\\
 \midrule
  UMDIR-GAN&1.202 (0.196) & 1.35 (0.14) & 0.067 (0.010) & 1.88 (0.32) & \bf{0.08}/3.60 \\
 UMDIR-Lat & {{1.146}} (0.232) &  1.04 (0.14) & {{0.067}} (0.012) & 1.37 (0.41) & \bf{0.08}/3.60\\

UMDIR-LaGAN & {\bf{1.126}} (0.214) & 0.97 (0.11) & {{0.064}} (0.010) & 1.05 (0.20) & \bf{0.08}/3.60\\
    \bottomrule
  \end{tabular}}
\end{table*}

\begin{figure}[!t]
\centering
\includegraphics[width=0.94\linewidth]{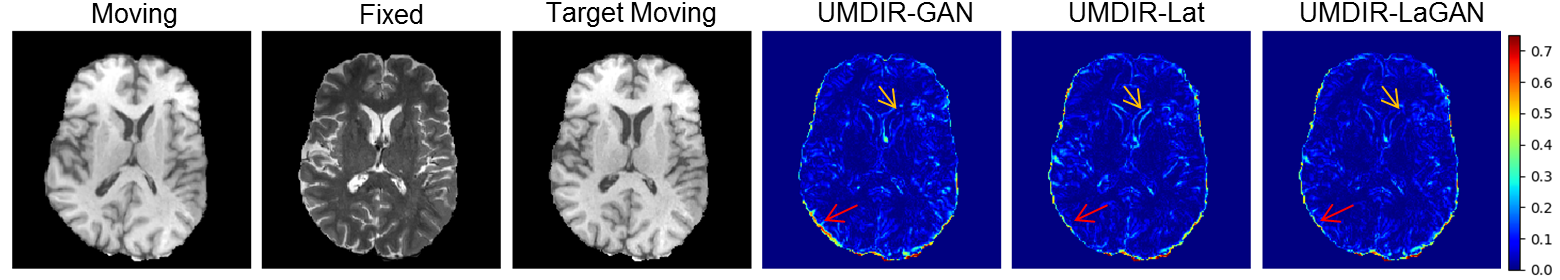}
\caption{Comparison between variants of proposed methods for T1 to T2 registration. Error maps are pixel-wise intensity differences between warped images and target moving images. Red arrows indicate regions where UMDIR-Lat outperforms UMDIR-GAN, and yellow arrows point out regions where UMDIR-GAN outperforms UMDIR-Lat. Combined metrics shows better performance than each of them separately.}
\label{fig:error}
\vspace{-4mm}
\end{figure}

Additionally, we also evaluated the bi-directional multi-modal registration performance on BraTS dataset, where ground truth deformation fields are available from T2 to T1, and ground truth aligned images are known for the T1 to T2 registration. Note that our proposed latent method can realize such bi-directional registration directly without any further training, while other competing methods need to be optimized or trained separately for both T1 $\rightarrow$ T2 and T2 $\rightarrow$ T1 directions. Quantitative results are shown in Table \ref{BraTS_table}, with RMSE($\Delta$) and Grad Det-Jac calculated for T2 $\rightarrow$ T1, and RMSE(I) for T1 $\rightarrow$ T2. An example of visualization results is shown in Fig. \ref{fig:COPD}. From both quantitative and qualitative results, it can be seen that the proposed UMDIR methods can achieve competitive registration performance compared with other competing methods and also with smooth and regular deformations. {{Though MIND achieved a lower RMSE(I) for T1 $\rightarrow$ T2 registration, this is at the cost of separate training for each direction. Our proposed methods deliver a one-shot bi-directional solution in a noticeably higher processing speed. In addition, learning registration directly from disentangled latent representations bypasses the image-to-image translation stage, which can potentially avoid problems caused by the image translation quality such as image noises or hallucinations that can lead to inaccurate registration. This possibly explains the better performance of our proposed methods against I2I with mono-modal registration methods.  }}
On the other hand, to examine the differences of latent similarity criterion and cross-cycle adversarial similarity metric as well as the benefits of combined metrics, in Fig.\ref{fig:error}, we compared the registration performance with different metrics in the inverse direction (T1 $\rightarrow$ T2) by visualizing the pixel-wise intensity RMSE. It can be observed that by combining these two complementary metrics, UMDIR-LaGAN is able to preserve the advantages of each metric and produces more accurate registration.

\section{Conclusion}
In this paper, we have presented a novel deep learning based model for fully unsupervised  multi-modal deformable image registration. The proposed models reduce the multi-modal registration problem to a mono-modal one via exploiting the disentangled latent embedding that is learned from an unpaired image-to-image translation framework.  For training the registration network, we proposed a distance loss in latent shape space and a cross-cycle adversarial loss defined in image space as similarity metrics. Experimental results showed improvements of our proposed models against other conventional approaches in terms of both accuracy and speed. 

For future work, we will further investigate the impact of joint training of MUNIT and UMDIR networks on registration accuracy. It could be helpful to refine the image-to-image translation network during the training of the registration network, so that the domain-invariant attribute can be better enforced to be equivalent to the shape that could be represented via geometrical transformations. Additionally, we will also extend the application of our model on real 3D volumes via patch-based methods for efficient training.
\\[12pt]
\noindent \textbf{Disclaimer:} This feature is based on research, and is not commercially available. Due to regulatory
reasons its future availability cannot be guaranteed.

%
%
%
{\small{
\bibliographystyle{splncs04}
\bibliography{ref}
}}%
\end{document}